\documentclass{article}

\usepackage[utf8]{inputenc}
\usepackage{authblk}
\usepackage{setspace}
\usepackage[margin=1.25in]{geometry}
\usepackage{graphicx}
\graphicspath{ {./figures/} }
\usepackage{subcaption}
\usepackage{amsmath}
\usepackage{xcolor}
\usepackage{algorithm}
\usepackage{algorithmic}
\usepackage{siunitx}
\usepackage{ntheorem}
\usepackage{hyperref}
\theoremseparator{:}

\newtheorem*{hyp*}{Hypothesis \protect\hypnumber}

\newcommand{\hypnumber}{}

\usepackage[style=nejm, 
citestyle=numeric-comp,
sorting=none]{biblatex}
\title{ELM: A Hybrid Ensemble of Language Models for Automated Tumor Group Classification in Population-Based Cancer Registries}

\author[1,2*]{Lovedeep Gondara}
\author[1]{Jonathan Simkin}
\author[1]{Shebnum Devji}
\author[3]{Gregory Arbour}
\author[3]{Raymond Ng}

\affil[1]{British Columbia Cancer Registry, Provincial Health Services Authority, Vancouver, Canada.}
\affil[2]{School of Population and Public Health, University of British Columbia, Vancouver, Canada.}
\affil[3]{The Data Science Institute, University of British Columbia, Vancouver, Canada.}
\affil[*]{Address correspondence to: lovedeep.gondara@ubc.ca}

\date{}

\begin{document}

\maketitle

%%%%%% Abstract %%%%%%

\newpage

\begin{abstract}

\textbf{Background:} Population-based cancer registries (PBCRs) manually extract data from unstructured pathology reports, a labor-intensive process where assigning reports to tumor groups can consume 900 person-hours annually for approximately 100,000 reports at a medium-sized registry. Current automated rule-based systems fail to handle the linguistic complexity of this classification task.

\textbf{Materials and Methods:} We present ELM (Ensemble of Language Models), a novel hybrid approach combining small, encoder only language models and large language models (LLMs). ELM employs an ensemble of six fine-tuned encoder only models: three analyzing the top portion and three analyzing the bottom portion of each report to maximize text coverage given token limits. A tumor group is assigned when at least five of six models agree; otherwise, an LLM arbitrates using a carefully curated prompt constrained to likely tumor groups.

\textbf{Results:} On a held-out test set of 2,058 pathology reports spanning 19 tumor groups, ELM achieves weighted precision and recall of 0.94, representing a statistically significant improvement ($p<0.001$) over encoder-only ensembles (0.91 F1-score) and substantially outperforming rule-based approaches. ELM demonstrates particular gains for challenging categories including leukemia (F1: 0.76→0.88), lymphoma (0.76→0.89), and skin cancer (0.44→0.58).

\textbf{Discussion:} Deployed in production at British Columbia Cancer Registry, ELM has reduced manual review requirements by approximately 60-70\%, saving an estimated 900 person-hours annually while maintaining data quality standards.

\textbf{Conclusion:} ELM represents the first successful deployment of a hybrid small, encoder only models-LLM architecture for tumor group classification in a real-world PBCR setting, demonstrating how strategic combination of language models can achieve both high accuracy and operational efficiency.

\textbf{Keywords:} cancer registries, ensemble learning, natural language processing, large language models, clinical text classification, pathology reports

\end{abstract}

\newpage
%%%%%% Main Text %%%%%%

\section{Introduction}
Globally, there are an estimated 10.4 million cancer deaths and 20.0 million new cancer cases in 2022, projected to increase to 18.5 million deaths and 35.3 million new cases by 2050\footnote{\url{https://gco.iarc.who.int}} \cite{sung2021global}. Population-Based Cancer Registries (PBCRs) serve the critical role of collecting standardized data on all cancer cases within defined populations \cite{parkin2006evolution}. The primary data source for PBCRs consists of pathology reports—unstructured clinical documents generated from pathologists' examination of tissue and fluid samples, containing essential information about diagnosis, cancer type, biomarkers, and disease stage.

Structured data extracted from pathology reports populate cancer registry databases that underpin cancer surveillance, prevention programs, and clinical research \cite{parkin2006evolution}. However, this data extraction represents a monumental challenge given the volume involved: medium-sized PBCRs process hundreds of thousands of pathology reports annually \cite{chtourou2023impact}. PBCRs currently rely on trained registrars and subject matter experts (SMEs) to manually review and abstract data from these complex clinical documents.

\subsection{The Tumor Group Classification Challenge}

Tumor group assignment—the process of categorizing cancers into broad anatomical and histological categories—represents a foundational step in cancer registry workflows. Accurate tumor group classification enables appropriate routing of cases to specialized registrars, facilitates subsequent detailed coding of histology and site, and directly impacts the quality of surveillance statistics and research data. However, the task presents significant clinical and linguistic challenges. Reports contain complex medical terminology requiring substantial domain expertise. Phrases like ``skin of left lateral breast'' require understanding that a skin biopsy was performed on breast tissue, not a breast cancer diagnosis. Distinguishing primary tumors from metastatic lesions demands sophisticated contextual interpretation. Report formats vary substantially across institutions and individual pathologists, and not all reports contain complete diagnostic conclusions. These complexities exceed the capabilities of traditional rule-based natural language processing systems.

\subsection{Limitations of Current Automated Approaches}

Most PBCRs in the USA and Canada receive pathology reports digitally, often formatted according to Health Level 7 (HL7) standards. To reduce manual review burden, many registries employ Electronic Mapping, Reporting, and Coding (eMaRC) \cite{emarcplus}, a rule-based text analytics tool developed by the US Centers for Disease Control and Prevention. eMaRC analyzes digital pathology reports and attempts to identify tumor information and assign tumor groups based on anatomical terms.

However, rule-based systems fundamentally struggle with the nuances of clinical language \cite{lopez2022natural,santos2022automatic}. They rely on pattern matching and fail to capture long-range dependencies or understand context. For example, given the phrase \emph{``skin left lateral breast squamous cell carcinoma''}, eMaRC may incorrectly classify this as breast cancer or fail to classify it at all due to multiple anatomical sites, when the correct tumor group is skin cancer.

At the British Columbia Cancer Registry (BCCR), eMaRC fails to assign a tumor group to approximately 40\% of reportable cancer cases. When it does make assignments, the error rate necessitates complete manual review for quality assurance, effectively rendering the automation redundant. Registrars must review all $\sim$90,000 annual reportable cases regardless of eMaRC's output, consuming approximately 800-900 person-hours annually for tumor group assignment alone\footnote{Based on operational experience at British Columbia Cancer Registry}, assuming roughly 30 seconds per report.

\subsection{The ELM Approach}

To address these limitations, we developed ELM (Ensemble of Language Models), a novel hybrid architecture that strategically combines the strengths of multiple language model types. Our method leverages:

\begin{enumerate}
    \item \textbf{Six fine-tuned small, encoder-only language models} working in ensemble: three processing the beginning of each report and three processing the end—to maximize coverage despite token limitations
    \item \textbf{Ensemble voting with high agreement threshold} to ensure confident predictions on straightforward cases
    \item \textbf{Large language model (LLM) arbitration} for ambiguous cases or historically challenging tumor groups, with carefully engineered prompts constraining outputs to likely options
\end{enumerate}

This hybrid strategy offers several key advantages. First, ensemble diversity combined with full report coverage captures tumor group information that may appear anywhere in the document. Second, computational efficiency is achieved by routing only uncertain cases (approximately 15-20\%) to the expensive LLM. Third, LLM behavior is controlled through constrained output spaces and structured response formats. Finally, the approach provides explainability through LLM-generated reasoning for each arbitrated case. We demonstrate that this approach achieves state-of-the-art performance while providing substantial operational efficiency gains in a real-world PBCR deployment.

\section{Methods}
We describe ELM in a general framework before detailing our specific implementation at BCCR.

\subsection{The Initial Ensemble}
\label{sec:ensemble}

The foundation of ELM consists of $X$ discriminative encoder-only language models ($ML_1,\cdots,ML_X$), each fine-tuned on labeled pathology reports to classify documents into predefined tumor groups. We provide implementation specifics in Section \ref{sec:hex_imp}.

\subsubsection{Token Segmentation Strategy}

A critical design decision involves how to handle pathology reports that exceed the input token limits of transformer-based models. Rather than using a sliding window or other segmentation strategies, we split reports into top and bottom segments. This decision is motivated by the typical structure of pathology reports, where the top section contains the ``Final Diagnosis'' or ``Diagnosis'' section in structured reports with primary tumor information typically stated explicitly, while the bottom section contains pathologist interpretations, comments, or impressions that provide diagnostic reasoning and often clarify ambiguous findings. Middle sections often contain procedural details, specimen descriptions, gross examination findings, and technical information that, while contextually important, are less determinative for tumor group classification.

Each model processes exactly 512 tokens (the maximum input length for BERT-based architectures), representing approximately 1-2 paragraphs of text. This length was selected based on model architectural constraints, computational efficiency considerations, and empirical observation that tumor group-relevant information tends to be concentrated rather than dispersed throughout reports.

Half the models in the ensemble analyze the first 512 tokens of each report; the other half analyze the last 512 tokens. This ensures maximum coverage of diagnostically relevant content. Each model independently predicts a tumor group, after which votes are summed across the $m$ possible tumor group categories to produce a vote distribution $[c_1,\cdots,c_m]$.

\subsubsection{Ensemble Decision Logic}

If the maximum vote count exceeds a pre-specified threshold $v$ and the predicted tumor group is not in a set of known hard-to-classify categories $G$, the ensemble's prediction is accepted. Otherwise, the case proceeds to LLM arbitration. This flexible design allows domain experts to designate specific tumor groups for LLM review based on inherent linguistic ambiguity (for example, distinguishing non-melanoma skin cancer from melanoma) or limited training data availability for rare tumor types.

\subsection{Arbitrating LLM}
\label{sec:llm_method}

Cases not meeting the ensemble confidence threshold, or classified into pre-specified challenging tumor groups, are routed to a large language model for final determination. The LLM receives both the pathology report and a carefully engineered prompt.

\subsubsection{Prompt Engineering Rationale}

The prompt was iteratively refined through collaboration with subject matter experts and incorporates several critical design elements. First, role specification explicitly positions the LLM as a ``specialized pathology assistant'' to prime domain-appropriate reasoning patterns. Second, rather than asking the LLM to choose from all 19 tumor groups, the prompt constrains selection to a small targeted set (typically 2-4 options) based on encoder-only ensemble predictions plus domain expert knowledge of likely alternatives. This constraint substantially reduces hallucination risk and improves accuracy.

Third, the prompt provides explicit guidance on challenging distinctions: leukemia versus lymphoma based on presentation site (bone marrow versus lymph nodes), melanoma versus non-melanoma skin cancer (explicit mention of melanoma takes precedence), cervical versus other gynecological cancers (cervix receives separate classification), and anatomical groupings (for example, gastrointestinal encompasses esophageal, gastric, pancreatic, anal, and small bowel tumors). Finally, the prompt mandates JSON format with both the predicted tumor group and reasoning, ensuring parseable output and providing explainability for quality assurance and model refinement.

We selected a locally-deployable LLM (Mistral Nemo Instruct-2407) over cloud-based alternatives to ensure compliance with healthcare privacy regulations while maintaining strong performance.

The complete ELM algorithm is formalized in Algorithm \ref{algo:overview}.

\begin{algorithm}[h]
\caption{Overview of ELM}\label{algo:overview}
\begin{algorithmic}[1]
\REQUIRE A set of small, encoder-only language models $X$, a single LLM $L$ for arbitration, observations $d_1,\cdots, d_n$ for inference, voting threshold $v$ for arbitration, number of tumor groups $m$, hard-to-classify tumor groups $\{g_1,\cdots,g_n\} \in G$
\FORALL {$d_i$, $i \in 1,\cdots, n$}
   \STATE Predict the tumor group for $d_i$ using all models in $X$
   \STATE Sum the votes per tumor group $[c_1,\cdots,c_m]$
   \STATE Store the predicted tumor groups in $t_i$
\IF {$max([c_1,\cdots,c_m]) < v$ $\mid$ {$t_i \in G$}}
   \STATE $L$ gets as input a prompt and $d_i$ to predict tumor group out of $[t_i, G]$
   \STATE Output the predicted tumor group by $L$
\ENDIF
\IF {$max([c_1,\cdots,c_m])  \ge v$ \& {$t_i \notin G$}}
\STATE output the tumor group with majority vote
\ENDIF
\ENDFOR
\end{algorithmic}
\end{algorithm}

\noindent\textbf{Note:} In our implementation, $v=5$ (requiring 5-of-6 agreement), and $G$ includes \{cervix, multiple myeloma, primary unknown, skin\} based on domain expert input identifying these as historically challenging classifications.

\section{Implementing ELM}
\label{sec:hex_imp}

\begin{figure*}[]
    \centering
    \includegraphics[scale=0.6]{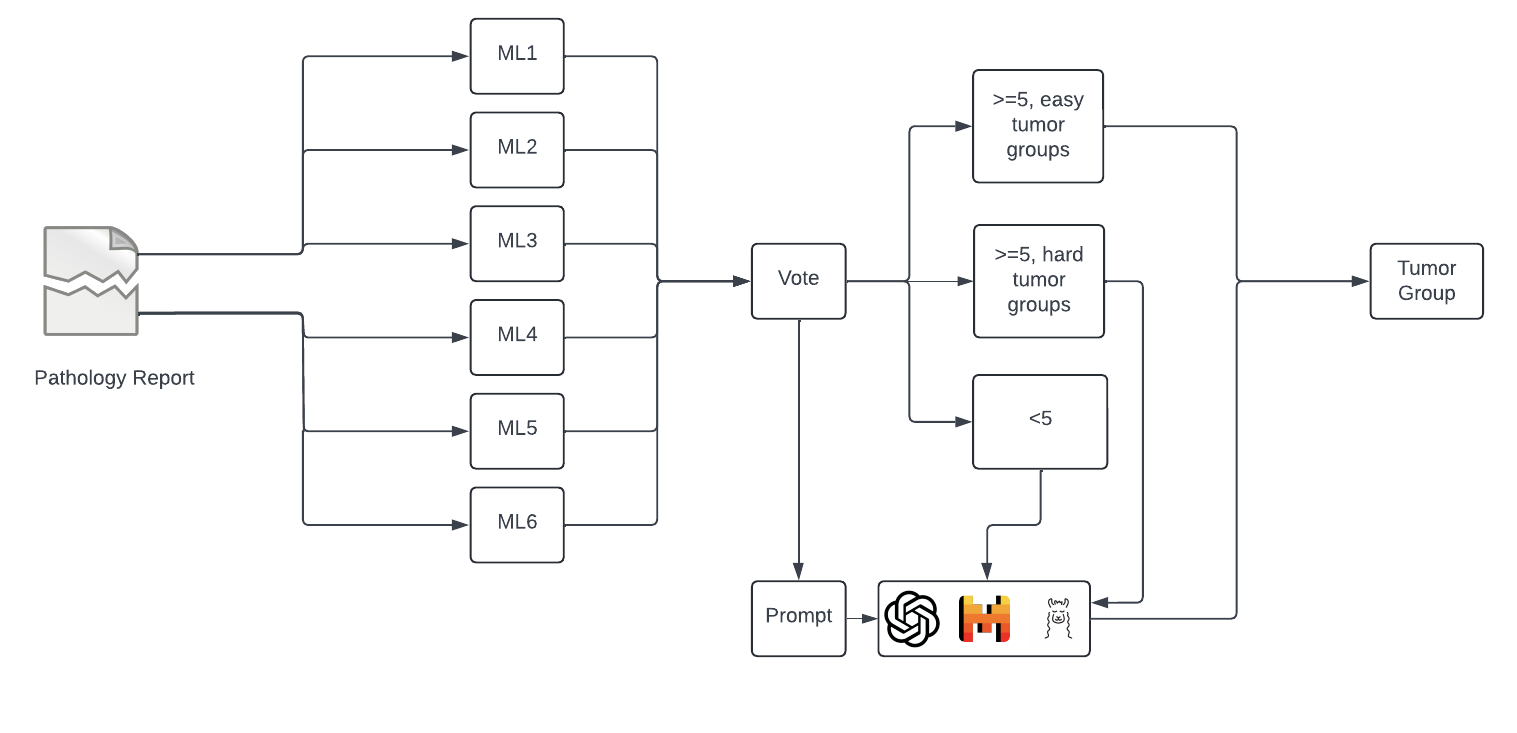}
    \caption{ELM in action. A pathology report is sent to six small, encoder-only language models for classification. After summing votes per tumor group, if the majority vote falls below the threshold (5 in this implementation) or the predicted tumor group belongs to historically difficult categories, the report proceeds to LLM arbitration. The LLM receives a carefully curated prompt that constrains selection to likely tumor groups based on encoder-only model predictions and domain expert knowledge.}
    \label{fig:dxdesign}
\end{figure*}

This section details our specific implementation at the British Columbia Cancer Registry. Figure \ref{fig:dxdesign} illustrates the complete ELM workflow.

\subsection{Small, Encoder-only Language Models}
\label{sec:slm}

We employ three distinct pre-trained clinical language models, each fine-tuned separately on top and bottom report segments, yielding six specialized models total. Fine-tuning used 16,000 annotated pathology reports spanning 19 tumor groups, trained for 3 epochs with learning rate $2e^{-5}$. Top models process the first 512 tokens; bottom models process the last 512 tokens. Table \ref{tab:model_det} summarizes the base models.

BCCRTron represents a further domain-specialized version of GatorTron \cite{yang2022gatortron}, additionally pre-trained on BCCR's pathology report corpus in an unsupervised manner\footnote{Model available upon request}.

\begin{table}[h]
\centering
\begin{tabular}{|l|l|}
\hline
\textbf{Base Model} & \textbf{Section} \\ \hline
GatorTron \cite{yang2022gatortron}  & Top \& Bottom           \\ \hline
BCCRTron  & Top \& Bottom           \\ \hline
ClinicalBERT \cite{alsentzer2019publicly}  & Top \& Bottom            \\ \hline
\end{tabular}
\caption{Small, encoder-only language models used in ELM. Each base model is fine-tuned separately on top and bottom report sections, producing six specialized classifiers.}
\label{tab:model_det}
\end{table}

\subsubsection{Computational Efficiency Considerations}

The hybrid architecture provides substantial computational advantages over LLM-only approaches. In production deployment, encoder-only ensemble inference requires approximately 600ms per report (six models at roughly 100ms each), while LLM inference requires 2-3 seconds per report when needed. Approximately 15-20\% of reports require LLM arbitration. This means ELM processes the majority of reports (80-85\%) in under one second using only the encoder-only ensemble, with average per-report processing time of approximately 850ms including cases requiring LLM arbitration. A pure LLM approach would require 2-3 seconds per report, representing a 3-4$\times$ increase in both computational cost and processing time. For a registry processing 90,000 reportable pathology reports annually, this efficiency translates to the difference between feasible real-time processing and prohibitive computational expenses.

\subsection{Large Language Model}

For LLM arbitration, we employ Mistral Nemo Instruct-2407 \cite{jiang2023mistral,jiang2024mixtral}, a 12-billion parameter instruction-tuned model. We use it in zero-shot mode with a domain-specific prompt (Appendix) designed to capture nuances of tumor group distinctions including leukemia vs. lymphoma, melanoma vs. non-melanoma skin cancer, and cervical vs. other gynecological tumors. The prompt was developed iteratively with input from cancer registry SMEs.

We also evaluated smaller alternatives—Llama 3.2 (3B parameters) \cite{touvron2023llama} and Qwen 2.5 (3B parameters) \cite{qwen2.5}—using identical prompting strategies. Comparative results appear in Section \ref{sec:results}.

\subsection{Data}

As of 2020, BCCR receives all provincial pathology reports as HL7 messages in real-time from regional laboratory information systems, totaling approximately 1 million reports annually (cancer and non-cancer combined). HL7 is a healthcare messaging standard \cite{eichelberg2005survey} defining structure and content for clinical document exchange. After exclusions (out-of-province patients, etc.), approximately 600,000 messages proceed to eMaRC \cite{emarcplus} processing.

eMaRC version 6.0.0.5 attempts to distinguish cancer from non-cancer pathology by searching for terms related to anatomical site, histology, behavior, grade, and laterality \cite{emarcplus}. Approximately 130,000 reports are flagged as potentially containing cancer information. These undergo further filtering using an SLM-based approach \cite{gondara2024classifying} to separate reportable ($\sim$90,000) from non-reportable ($\sim$40,000) cancer cases.

The next critical step involves assigning tumor groups to reportable cases. Here, eMaRC's limitations become apparent: it fails to assign any tumor group label to approximately 40\% of cases, and for cases receiving assignments, error rates necessitate complete manual review. This operational reality motivated ELM's development.

\subsubsection{Training Data}

Our training set comprises 16,000 post-eMaRC reportable pathology reports from 2020, each labeled with one of 19 tumor groups used at BCCR (breast, colorectal, lung, skin, melanoma, lymphoma, leukemia, etc.). Labels were assigned at the patient level by expert registrars based on final cancer diagnoses.

\textbf{Label Quality Considerations:} Patient-level labeling introduces controlled label noise at the individual report level. A single patient may have multiple pathology reports from different anatomical sites. For example, a patient with a final prostate cancer diagnosis may also have biopsy samples from skin or other genitourinary sites; all reports for this patient receive the ``prostate'' label despite some potentially describing incidental findings from other sites.

This patient-level labeling approach is acceptable for several reasons. For most patients (over 85\%), all pathology reports genuinely pertain to the same tumor group, so patient-level labels are correct. BERT-based transformer models demonstrate robustness to label noise in the 10-15\% range \cite{agro2023handling}. Our large training set (n=16,000) provides sufficient signal despite noise. Most importantly, our test set uses clean report-level labels for unbiased evaluation. We estimated training label noise at approximately 10-15\% through manual audit of 200 randomly selected reports. Detailed training set sizes per tumor group appear in the Appendix.

\subsubsection{Test Data}

To ensure unbiased evaluation, we constructed a test set of 2,058 pathology reports randomly selected from 2023-2024 data. Each report was individually labeled by expert cancer registrars specifically for this evaluation, ensuring report-level (not patient-level) ground truth labels. These reports were completely unseen during model training. Test set size per tumor group is detailed in the Appendix.

\section{Results}
\label{sec:results}

We evaluate ELM using weighted average precision, recall, and F1-score across 19 tumor groups. Weighted averaging accounts for class imbalance in the test set.

\subsection{Baseline Comparisons}

Table \ref{tab:baseline_comp} presents performance across multiple approaches, demonstrating ELM's superiority. eMaRC performance estimates derive from operational data at BCCR. The LLM-only approach uses Mistral Nemo with a prompt listing all 19 tumor groups (no ensemble guidance).

\begin{table}[h]
\centering
\begin{tabular}{|l|c|c|c|p{3.5cm}|}
\hline
\textbf{Approach} & \textbf{Precision} & \textbf{Recall} & \textbf{F1} & \textbf{Notes} \\ \hline
eMaRC (rule-based) & 0.71 & 0.58 & 0.64 & 40\% no prediction \\ \hline
Single encoder only (avg) & 0.91 & 0.86 & 0.87 & First or last 512 tokens only \\ \hline
Encoder only Ensemble & 0.93 & 0.91 & 0.91 & No LLM arbitration \\ \hline
LLM-only (Mistral) & 0.89 & 0.87 & 0.88 & All 19 groups, no ensemble guidance \\ \hline
\textbf{ELM (proposed)} & \textbf{0.94} & \textbf{0.94} & \textbf{0.94} & \textbf{Hybrid approach} \\ \hline
\end{tabular}
\caption{Performance comparison across baseline approaches. ELM achieves the highest performance across all metrics. eMaRC metrics estimated from BCCR operational data; all others evaluated on the held-out test set (n=2,058).}
\label{tab:baseline_comp}
\end{table}

\subsection{The Need for an Ensemble}

We first justify the ensemble approach by comparing individual encoder-only model's performance against the six-model ensemble (without LLM arbitration). For single-model metrics, we report the average across all six models. Table \ref{tab:sing_mult} demonstrates substantial gains from ensembling.

Performance improvements stem from two factors: (\emph{i}) ensemble diversity reduces individual model errors, and (\emph{ii}) analyzing both top and bottom report sections captures diagnostic information that may appear anywhere in the document, versus the limited 512-token window of individual models.

\begin{table}[h]
\centering
\begin{tabular}{|l|l|l|}
\hline
\textbf{Metric (Avg.)}    & \textbf{Single Model} & \textbf{Ensemble} \\ \hline
Precision & 0.91         & 0.93         \\ \hline
Recall    & 0.86         & 0.91         \\ \hline
F1-Score  & 0.87         & 0.91         \\ \hline
\end{tabular}
\caption{Comparing average single-model performance to the six-model ensemble (without LLM).}
\label{tab:sing_mult}
\end{table}

\subsection{The Need for LLM Arbitration}

Having established ensemble benefits, we now examine the value of LLM arbitration. When fewer than five of six models agree, or when the predicted tumor group belongs to $G$ (hard-to-classify categories), the case proceeds to the LLM with a constrained prompt.

Table \ref{tab:hexllm} shows that LLM arbitration yields statistically significant improvement (McNemar's test, $p < 0.001$). This gain reflects the LLM's superior ability to handle nuanced language and make fine-grained distinctions when properly guided by the ensemble and constrained prompt.

\begin{table}[h]
\centering
\begin{tabular}{|l|l|l|}
\hline
\textbf{Metric (Avg.)}    & \textbf{Encoder only Ensemble} & \textbf{ELM} \\ \hline
Precision & 0.93        & 0.94   \\ \hline
Recall    & 0.91        & 0.94   \\ \hline
F1-Score  & 0.91        & 0.94   \\ \hline
\end{tabular}
\caption{The importance of LLM arbitration in ELM. The improvement is statistically significant (McNemar's test, p<0.001).}
\label{tab:hexllm}
\end{table}

\subsection{Impact of LLM Size}

We evaluated whether smaller LLMs could reduce computational requirements while maintaining acceptable performance. Table \ref{tab:sllm} compares Mistral Nemo (12B parameters) against Llama 3.2 and Qwen 2.5 (both 3B parameters), with all other settings held constant.

Results indicate that the 12B model provides the best performance, though smaller LLMs still outperform the encoder-only ensemble (F1=0.91). For registries with severe computational constraints, 3B models represent viable alternatives, accepting a modest performance reduction (2-3 percentage points) for substantially lower inference costs.

\begin{table}[h]
\centering
\begin{tabular}{|l|l|l|l|}
\hline
\textbf{Metric (Avg.)}    & \textbf{ELM (Mistral)} & \textbf{ELM (Llama)} & \textbf{ELM (Qwen)} \\ \hline
Precision & 0.94        & 0.93 & 0.93   \\ \hline
Recall    & 0.94        & 0.92 & 0.92   \\ \hline
F1-Score  & 0.94        & 0.92 & 0.92   \\ \hline
\end{tabular}
\caption{ELM performance with different LLM sizes. Smaller 3B parameter models offer reduced computational cost at modest performance reduction.}
\label{tab:sllm}
\end{table}

\subsection{Impact on Individual Tumor Groups}

Some tumor groups present inherent classification challenges due to linguistic ambiguity or limited training data (rare cancers). LLM arbitration particularly benefits these categories. Notable improvements include leukemia (F1 increased from 0.76 to 0.88, a 16\% improvement), lymphoma (F1 increased from 0.76 to 0.89, a 17\% improvement), sarcoma (F1 increased from 0.81 to 0.85, a 5\% improvement), and head and neck (F1 increased from 0.83 to 0.87, a 5\% improvement). Complete per-tumor-group results, including training and test set sizes, appear in Appendix Table A1.

\subsection{Strategic Routing of Hard-to-Classify Groups}

Beyond using the voting threshold to trigger LLM arbitration, ELM allows explicit designation of tumor groups known to be problematic. We configured ELM to route cervix, multiple myeloma, primary unknown, and skin tumor groups to the LLM regardless of ensemble confidence.

This strategic routing yielded substantial gains for several challenging categories. For cervix, F1 improved from 0.90 to 0.98 (a 9\% improvement). Multiple myeloma improved from 0.87 to 0.91 (a 5\% improvement). Non-melanoma skin cancer showed particularly dramatic improvement from 0.58 to 0.74 (a 28\% improvement). Primary unknown, an inherently ambiguous category, showed modest improvement from 0.38 to 0.40. This demonstrates that domain knowledge about challenging categories can be directly encoded into the system architecture for maximum benefit.

\subsection{Error Analysis}

We analyzed the 123 cases (6\% of test set) where ELM made incorrect predictions. The most common error pattern involved metastatic disease ambiguity (31\% of errors), where reports describing biopsies of metastatic lesions could be confused with primary tumors. For example, a report stating ``Liver biopsy showing adenocarcinoma consistent with colorectal primary'' was predicted as gastrointestinal instead of colorectal. 

Rare tumor types accounted for 23\% of errors, particularly tumor groups with severely limited training data such as ophthalmic  (n=142 training cases) and primary unknown (n=461 training cases). Multi-primary cancers represented 19\% of errors, where reports documented multiple synchronous primaries and single tumor group assignment was genuinely ambiguous. 

Ambiguous anatomical descriptions caused 15\% of errors, involving terminology that could span multiple tumor groups. For instance, ``nasopharyngeal mass'' could indicate head and neck cancer or potentially involve brain structures. Finally, incomplete reports accounted for 12\% of errors, where reports lacked final diagnosis sections due to pending ancillary tests or inadequate specimens.

These error patterns suggest opportunities for improvement through augmented training data for rare tumor types via active learning, multi-label classification architectures for multi-primary cases, and enhanced prompt engineering specifically targeting metastatic disease disambiguation.

\section{Discussion}

To our knowledge, ELM represents the first study to achieve state-of-the-art tumor group classification in a PBCR setting through a hybrid architecture combining small, encoder-only and large language models. This contrasts sharply with rule-based NLP approaches that dominate current PBCR practice. Rule-based systems like eMaRC rely on pattern matching and expert-crafted rules, making them brittle in the face of linguistic variation, unable to capture long-range dependencies, and prone to errors when reports mention multiple anatomical sites or contain complex clinical reasoning \cite{santos2022automatic,hammami2021automated,emarcplus,osborne2016efficient}.

Similarly, traditional machine learning approaches based on bag-of-words or term frequency models ignore crucial sentence-level representations and contextual information \cite{santos2022automatic}. For complex tasks like tumor group classification from lengthy, variable-format pathology reports, capturing sentence representations and broader context is essential—strengths of transformer-based architectures \cite{vaswani2017attention}.

Many transformer variants have been developed for clinical domains: BioBERT \cite{lee2020biobert}, ClinicalBERT \cite{alsentzer2019publicly}, PubMedBERT \cite{pubmedbert}, PathologyBERT \cite{santos2022automatic}, CancerBERT \cite{zhou2022cancerbert}, Path-BigBird \cite{chandrashekar2024path}, GatorTron \cite{yang2022gatortron}, and BiomedRoBERTa \cite{gu2021domain}. However, none of these models, nor any published studies employing them, have specifically addressed tumor group classification in real-world PBCR operations using pathology reports.

\subsection{Clinical Impact and Operational Validation}

Beyond technical performance metrics, ELM demonstrates tangible operational impact. Prior to deployment, tumor group assignment at BCCR required manual review of all $\sim$90,000 reportable pathology reports annually, consuming approximately 900 person-hours. With ELM in production, registrars now focus manual review on the 15-20\% of cases flagged by the ensemble for LLM arbitration, cases classified into historically challenging tumor groups, and a random 5\% sample for ongoing quality monitoring.

This represents a 60-70\% reduction in manual review workload, translating to substantial time savings that registrars can redirect toward more complex cognitive tasks requiring human expertise: detailed cancer staging, reconciliation of multiple primaries, ambiguous histology resolution, and patient follow-up data abstraction. Critically, post-deployment quality audits conducted quarterly show no increase in tumor group assignment errors compared to fully manual processes, confirming that automation maintains data quality standards.

A potential concern when first considering LLM-only approaches is why not simply route all reports to the LLM? We empirically evaluated this strategy and found it performs significantly worse than ELM (Table \ref{tab:baseline_comp}: F1=0.88 vs. 0.94). When the LLM must choose from all 19 tumor groups without ensemble guidance, the ambiguous and variable language in pathology reports leads to more frequent errors. By contrast, ELM's constrained prompts limit LLM selection to 2-4 likely options based on ensemble votes and domain expertise, substantially improving accuracy while also providing computational efficiency.

\subsection{Generalizability and Practical Considerations}

The methodology presented here holds significant implications for PBCRs worldwide as they transition to electronic pathology reporting systems. Tumor group assignment represents a fundamental, resource-intensive step in registry workflows. Scaled across medium to large registries processing 50,000-200,000 reportable pathology reports annually, ELM has potential to save thousands of person-hours across the cancer surveillance ecosystem.

This study's strengths include training and evaluation on real-world population-based pathology data at BCCR, temporal separation between training (2020) and test (2023-2024) data to assess robustness, use of open-source base models facilitating replication, and demonstration of production deployment with validated operational impact.

\subsection{Limitations}

Several limitations warrant acknowledgment. First, while training and test data span different time periods, both originate from BCCR. External validation on data from other North American or international registries is needed to assess generalizability across regional variations in pathology reporting practices, terminologies, and workflows. Second, our approach has been validated exclusively on English-language reports from Canadian pathologists; performance on reports in other languages or from substantially different healthcare systems remains unknown and would likely require language-specific model selection and retraining. 

Third, like all ML systems, ELM requires continuous monitoring for data drift and periodic retraining as pathology reporting practices evolve. Our deployment includes quarterly audit procedures to detect performance degradation, but long-term maintenance burden should not be underestimated. Fourth, while far more efficient than LLM-only approaches, ELM still requires substantial computational resources including GPU access for real-time inference. Smaller registries or those in resource-constrained settings may face infrastructure barriers to adoption without cloud computing options. 

Fifth, performance remains suboptimal for rare tumor types such as primary unknown (F1=0.40) and ophthalmic  cancers (n=1 test case), as well as for inherently ambiguous categories. These cases may always require human expertise and clinical judgment that exceeds current AI capabilities. Finally, while ELM's LLM component generates reasoning for its decisions, the encoder-only ensemble operates as a relative black box. For cases requiring deep investigation or when predictions contradict registrar expectations, full manual review of the original report remains necessary.

\subsection{Future Directions}

Several promising avenues emerge for future work. Multi-registry validation through collaboration with SEER registries in the United States and international registries would validate ELM's generalizability and identify institution-specific fine-tuning needs. Extending ELM to handle reports documenting multiple synchronous primary cancers through multi-label classification architectures would address cases where single-label assignments are genuinely ambiguous. 

Implementing active learning strategies could continuously improve performance on rare tumor types by intelligently selecting high-uncertainty cases from underrepresented categories for expert review and model updating. Exploring architectures that jointly predict tumor group alongside more detailed site and histology codes could potentially improve all tasks through multi-task learning and shared representations. 

Systematic evaluation of advanced prompting strategies including few-shot learning, chain-of-thought reasoning, and self-consistency methods may further enhance LLM arbitration performance. Investigating recently released models in the 3-7B parameter range could identify options offering better cost-performance trade-offs than Mistral Nemo while maintaining accuracy. Finally, developing attention visualization and saliency mapping tools would help registrars understand which report sections drove model predictions, supporting trust, error debugging, and continuous learning.

\section{Conclusion}

Accurate and efficient classification of tumor groups from unstructured pathology reports remains a significant challenge for population-based cancer registries, traditionally requiring extensive manual review and substantial human resources. Current automated methods, particularly rule-based systems like eMaRC, struggle with the complexities and linguistic nuances of clinical text, resulting in high error rates and continued dependence on manual validation.

This paper introduced ELM, a novel hybrid ensemble architecture specifically designed to address this challenge within real-world PBCR operations. ELM strategically combines the strengths of multiple specialized small, encoder-only language models to achieve broad report coverage and robust initial classification, augmented by the advanced reasoning capabilities of a large language model for targeted arbitration of ambiguous or inherently difficult cases.

Our comprehensive evaluation on 2,058 held-out pathology reports spanning 19 tumor groups demonstrates the effectiveness of this hybrid strategy. The encoder-only ensemble significantly outperforms individual models, and LLM arbitration provides statistically significant further gains (F1-score: 0.94), particularly for historically challenging categories including leukemia, lymphoma, and skin cancers. We confirmed superior performance of a 12B parameter LLM compared to smaller 3B alternatives, while demonstrating ELM's flexibility to strategically route known difficult-to-classify tumor groups for LLM review, yielding substantial additional performance improvements.

By successfully automating a critical, labor-intensive step in the PBCR workflow, ELM delivers a practical, high-performing solution for cancer surveillance infrastructure. Deployed in production at British Columbia Cancer Registry, ELM has reduced manual review requirements by 60-70\%, saving approximately 900 person-hours annually—time that expert staff can redirect toward higher-level abstraction tasks requiring nuanced clinical judgment. This work demonstrates that carefully designed hybrid architectures, combining complementary strengths of different language model types with domain expertise, can achieve both technical excellence and meaningful operational impact in healthcare informatics applications. The success at BCCR underscores ELM's potential for adaptation and implementation by other registries worldwide seeking to modernize their data processing pipelines while maintaining the high data quality standards essential for cancer surveillance and research.

\section{List of Abbreviations}
\begin{description}
    \item BCCR: British Columbia Cancer Registry
    \item ELM: Ensemble of Language Models  
    \item eMaRC: Electronic Mapping, Reporting, and Coding
    \item HL7: Health Level 7
    \item JSON: JavaScript Object Notation
    \item LLM: Large Language Model
    \item ML: Machine Learning
    \item NLP: Natural Language Processing
    \item PBCR: Population Based Cancer Registry
    \item SME: Subject Matter Expert
\end{description}

\printbibliography

\newpage
\section*{Appendix}

\subsection*{An Example Prompt}

\emph{``You are a specialized pathology assistant that reads the pathology reports and outputs the tumor group from one of the suggested ones. Note that for leukemia and lymphoma, if the disease presents in the lymph nodes, then it is lymphoma, and if the disease presents in the bone marrow, then it is leukemia. The tumor group of gastrointestinal cancers includes anal, esophageal, gastric, pancreatic, and small bowel cancers. Colorectal cancers includes colon and rectum. For the tumor group of gynecological cancers, it includes endometrial, fallopian tube, ovarian, uterine, vaginal, and vulvar cancers. Cervical cancer should be coded as the tumor group of cervix. If the pathology report mentions melanoma, then the tumor group should be melanoma, even if there is a mention of skin. You also provide reasoning to your choice of tumor group. Respond only with valid JSON with the following fields 'tumor\_group' and 'reason'. Do not write an introduction or summary. Assign a tumor group to the following pathology report.''}

\subsection*{Detailed Tumor Group Results}

Table A1 presents complete per-tumor-group performance metrics, including training and test dataset sizes. These results demonstrate ELM's particular value for challenging categories with limited training data or inherent linguistic ambiguity.

\setcounter{table}{0}
\renewcommand{\thetable}{A\arabic{table}}

\begin{table}[h]
\centering
\begin{tabular}{|l|l|l|l|}
\hline
\textbf{Tumor Group} & \textbf{Train(n)/Test(n)} & \textbf{SLM Only} & \textbf{ELM} \\ \hline
Breast & 966/406 & 0.99 & 1.00 \\ \hline
Colorectal & 1047/197 & 0.94 & 0.95 \\ \hline
Cervix & 812/26 & 0.90 & 0.90 \\ \hline
Gastrointestinal & 1037/143 & 0.83 & 0.87 \\ \hline
Genitourinary & 1000/147 & 0.96 & 0.96 \\ \hline
Gynaecological & 1012/129 & 0.93 & 0.94 \\ \hline
Head and Neck & 1028/78 & 0.83 & 0.87 \\ \hline
Leukemia & 281/88 & 0.76 & 0.88 \\ \hline
Lung & 1025/218 & 0.93 & 0.95 \\ \hline
Lymphoma & 642/129 & 0.76 & 0.89 \\ \hline
Melanoma & 335/122 & 0.98 & 0.98 \\ \hline
Multiple Myeloma & 1023/24 & 0.87 & 0.87 \\ \hline
Neuroendocrine & 1029/43 & 0.94 & 0.95 \\ \hline
ophthalmic  & 142/1 & 0.67 & 1.00 \\ \hline
Prostate & 1121/201 & 0.98 & 0.98 \\ \hline
Primary Unknown & 461/16 & 0.32 & 0.38 \\ \hline
Sarcoma & 1041/37 & 0.81 & 0.85 \\ \hline
Skin & 1001/19 & 0.44 & 0.58 \\ \hline
Thyroid & 1000/34 & 1.00 & 1.00 \\ \hline
\end{tabular}
\caption{Detailed F1-scores per tumor group with training and test set sizes. Notable improvements from LLM arbitration include leukemia, lymphoma, head and neck, gastrointestinal, and particularly skin cancer (non-melanoma).}
\label{tab:det}
\end{table}

\newpage
\section*{Declarations}

\subsection*{Ethics approval and consent to participate}
This study is operational in nature, not clinical research. The need for informed consent was waived by the University of British Columbia - BC Cancer Research Ethics Board (REB). Use of data for this study was approved by UBC-BC Cancer REB (REB number H23-01445). This study adheres to the Declaration of Helsinki.

\subsection*{Consent for publication}
Not applicable.

\subsection*{Availability of data and materials}
This study uses identifiable patient data that cannot be publicly shared. De-identified model code and synthetic example data may be available upon reasonable request. For inquiries, contact the corresponding author (lovedeep.gondara@phsa.ca).

\subsection*{Competing interests}
The authors declare no competing interests.

\subsection*{Funding}
The authors received no external financial support for this research.

\subsection*{Authors' contributions}
LG conceived and designed the study, developed the methodology, conducted experiments, and drafted the manuscript. SD contributed to LLM prompt design and data preparation. JS contributed to study design, result interpretation, and manuscript drafting. GA and RN provided methodological guidance and result interpretation. All authors read, revised, and approved the final manuscript.

\subsection*{Acknowledgements}
The authors thank the cancer registrars at British Columbia Cancer Registry for their expertise in labeling test data and providing domain knowledge essential to this work.

\end{document}